\documentclass[conference]{IEEEtran}
\usepackage{cite}
\usepackage{amsmath,amssymb,amsfonts}
\usepackage{algorithmic}
\usepackage{graphicx}
\usepackage{textcomp}
\usepackage[table,xcdraw]{xcolor}
\def\BibTeX{{\rm B\kern-.05em{\sc i\kern-.025em b}\kern-.08em
    T\kern-.1667em\lower.7ex\hbox{E}\kern-.125emX}}
\begin{document}

\title{Deep Reinforcement Learning for Foreign Exchange Trading\\
}

\author{\IEEEauthorblockN{1\textsuperscript{st} Yun-Cheng Tsai}
\IEEEauthorblockA{\textit{School of Big Data Management} \\
\textit{Soochow University}\\
Taipei, Taiwan \\
pecutsai@gm.scu.edu.tw}
\and
\IEEEauthorblockN{2\textsuperscript{nd} Chun-Chieh Wang}
\IEEEauthorblockA{\textit{Department of Computer Science}\\
{and Information Engineering} \\
\textit{National Taipei University}\\
Taipei, Taiwan}
}

\maketitle

The corresponding author is supported in part by the Ministry of Science and Technology of Taiwan under grant 107-2218-E-002-031. The first author is supported in part by the Ministry of Science and Technology of Taiwan under grant 107-2218-E-002-065.

\begin{abstract}
Reinforcement learning can interact with the environment and is suitable for applications in decision control systems. Therefore, we used the reinforcement learning method to establish a foreign exchange transaction, avoiding the long-standing problem of unstable trends in deep learning predictions. In the system design, we optimized the Sure-Fire statistical arbitrage policy, set three different actions, encoded the continuous price over a period of time into a heat-map view of the Gramian Angular Field (GAF) and compared the Deep Q Learning (DQN) and Proximal Policy Optimization (PPO) algorithms. To test feasibility, we analysed three currency pairs, namely EUR/USD, GBP/USD, and AUD/USD. We trained the data in units of four hours from 1 August 2018 to 30 November 2018 and tested model performance using data between 1 December 2018 and 31 December 2018. The test results of the various models indicated that favourable investment performance is achieved as long as the model is able to handle complex and random processes and the state is able to describe the environment, validating the feasibility of reinforcement learning in the development of trading strategies.
\end{abstract}


\section{Introduction}
We plan to use deep-enhanced learning to mimic how humans make decisions, using the state of the current environment to execute actions and obtain rewards from the environment. Moreover, people's actions impact the environment, causing the situation to enter a new state. To check the feasibility of this approach, we adopted the method of training four-hour units of EUR/USD, GBP/USD, and AUD/USD data between 1 August 2018 and 30 November 2018. We then applied the trained data to the period between 1 December 2018 and 31 December 2018 to validate the system performance.

\section{Preliminary}
In the paper, we adopted the Sure-Fire arbitrage strategy, which is a variant of the Martingale. It involves increasing bets after every loss so that the first win recovers all previous losses plus a small profit. After entering the market and initiating trading, the investor uses the margin between the stop-loss and stop-gain prices as the raised margin. As long as the price fluctuates within the increased margin and touches on the risen price, the Sure-Fire Strategy urges investors to continue growing the stakes until they surpass the margin to profit.
\begin{figure}[htbp]
\centering
\begin{minipage}[t]{0.3\textwidth}
\centering
\centerline{\includegraphics[width=5cm]{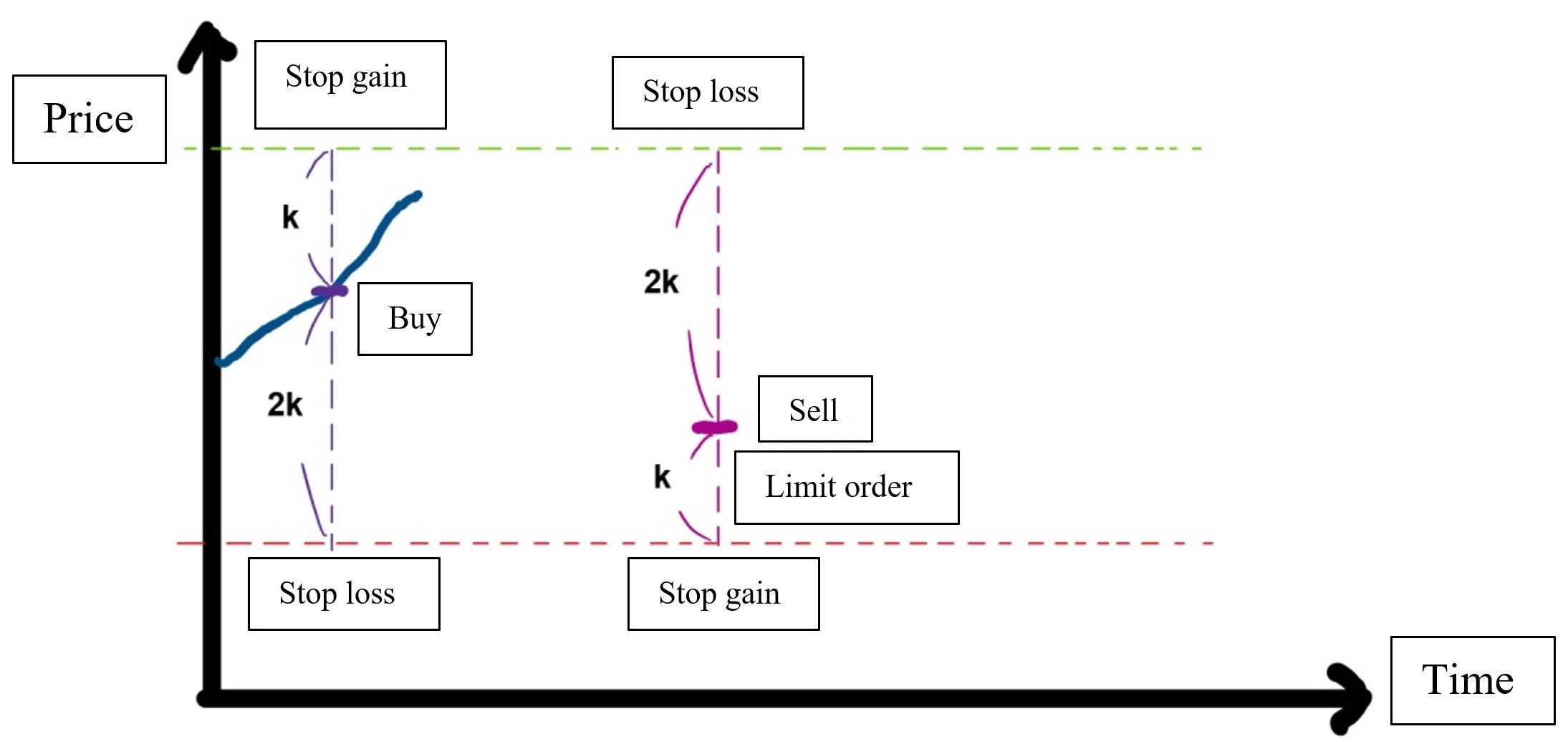}}
\caption{Sure-Fire arbitrage strategy 1.}\label{fig2}
\end{minipage}
\begin{minipage}[t]{0.3\textwidth}
\centering
\centerline{\includegraphics[width=5cm]{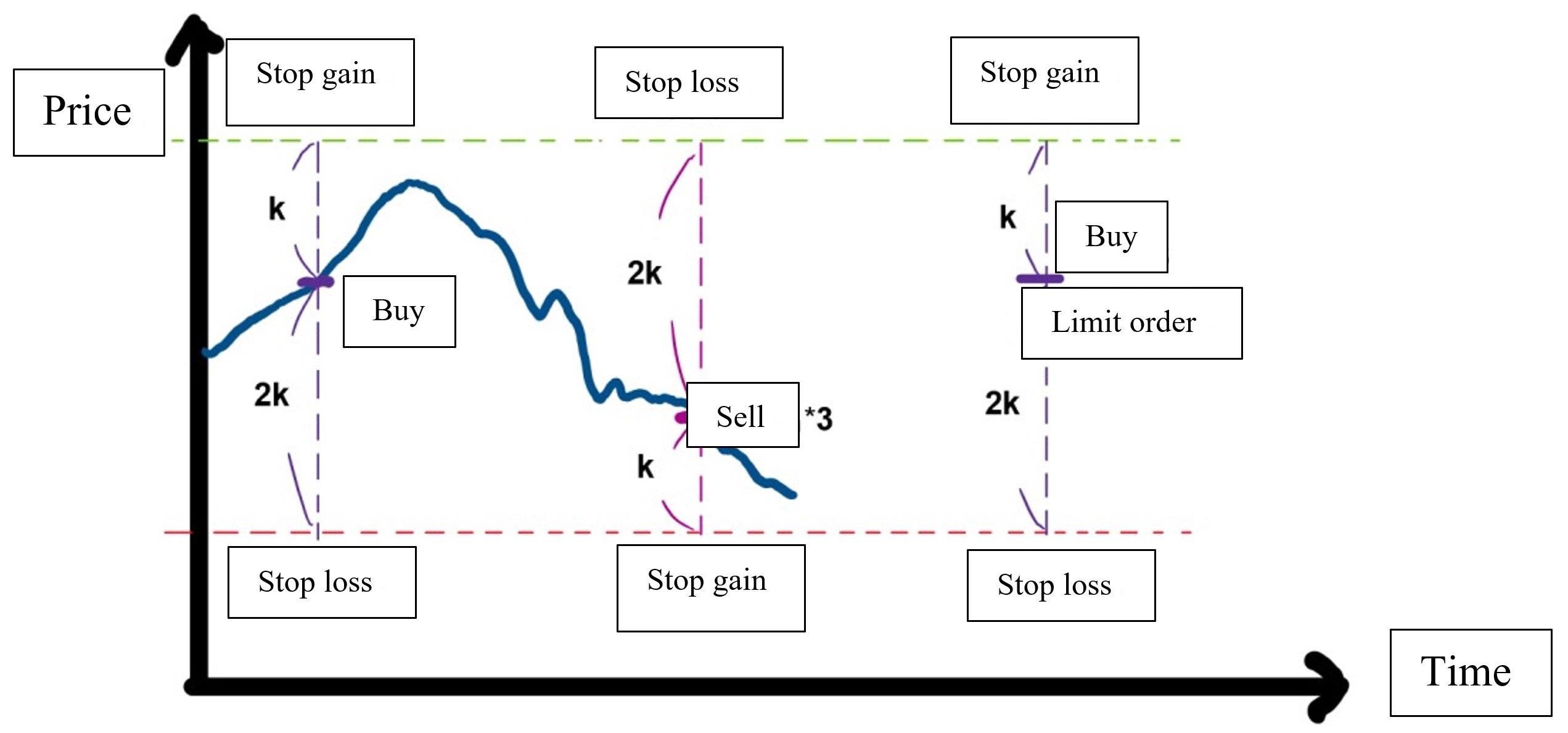}}
\caption{Sure-Fire arbitrage strategy 2.}\label{fig3}
\end{minipage}
\begin{minipage}[t]{0.36\textwidth}
\centering
\centerline{\includegraphics[width=6cm]{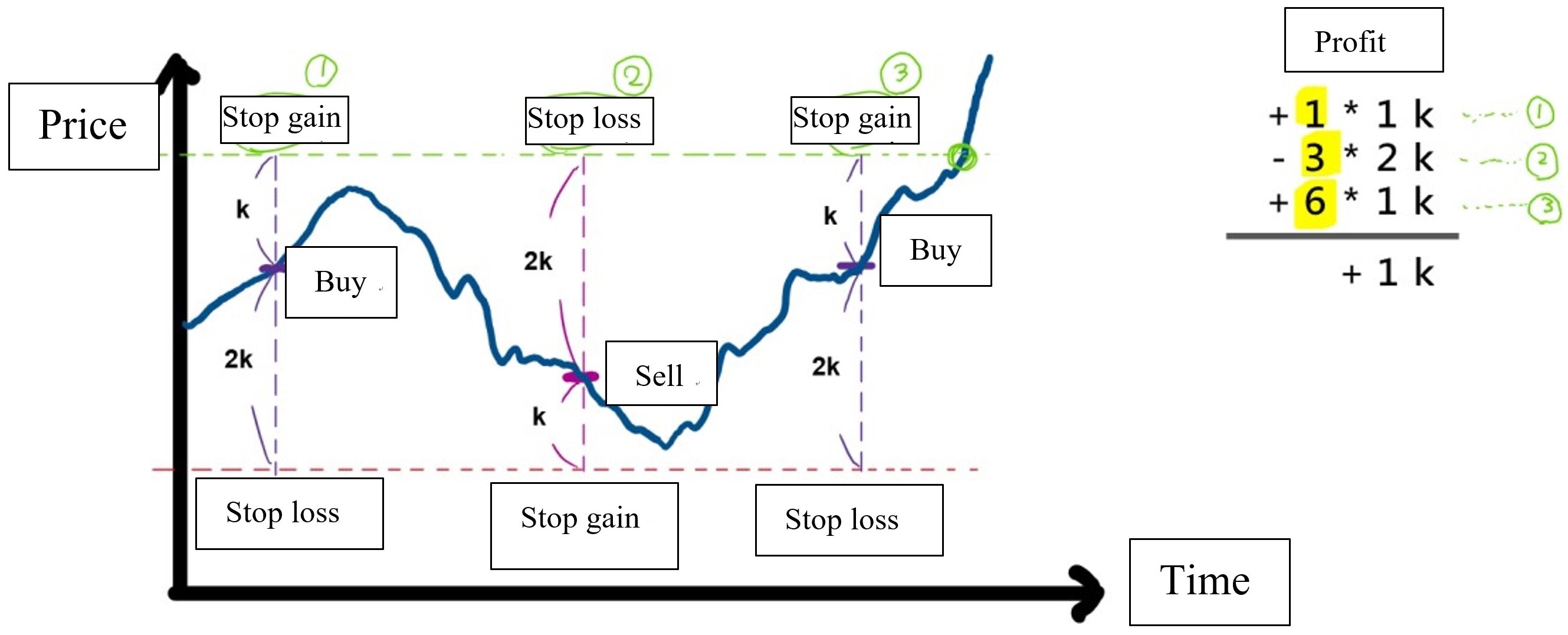}}
\caption{Sure-Fire arbitrage strategy 3.}\label{fig4}
\end{minipage}
\end{figure}
The most prominent shortcoming of Martingale is the lack of stakes or funds when the favorable odds are low. Therefore, we applied reinforcement learning to optimize the Sure-Fire Strategy. Data train to obtain the trading behavior with the minimum number of raises to achieve the maximum winning odds. A detailed transactions are illustrated from Fig.~\ref{fig2} to Fig.~\ref{fig4}. In Sure-Fire arbitrage strategy 1, we purchase one unit at any price and set a stop-gain price of $+k$ and a stop-loss price of $-2k$. At the same time, we select an amount with a difference of $-k$ to the buy price and $+k$ to the stop-loss price and set a backhand limit order for three units. Backhand refers to engaging in the opposite behavior. The backhand of buying is selling, and the backhand of selling is buying. A limit order refers to the automatic acquisition of corresponding units. In Sure-Fire arbitrage strategy 2, we place an additional backhand limit order, where the buy price is $+k$ to the selling price, and $-k$ to the stop-loss price when a limit order is triggered, and three units successfully sold backhand. We set the stop-gain point as the difference of $+k$ and the stop-loss point as the difference of $-2k$, after which an additional six units buy. In Sure-Fire arbitrage strategy 3, the limit order triggers in the third transaction. The final price exceeded the stop-gain amount of the first transaction, the stop-loss price of the second transaction, and the stop-gain price of the third transaction. In this instance, the purchase is complete. The calculation in the right block shows that the profit is $+1k$.

\subsection{Deep Q Network (DQN)}
The Deep Q Network (DQN) is a deep, reinforcement learning framework~\cite{hausknecht2015deep}. For the selection of actions, if the greedy method alone is applied to select the action with the highest expected return every round, then the chance of selecting other actions would be lost. Therefore, DQN adopts the $\epsilon$-greedy exploration method. Specifically, each time an action is selected, the system provides a small probability ($\epsilon$) for exploration, in which the best action abandoned and other new actions executed. The probability of exploration increases concurrently with the $\epsilon$ value, making the agent more likely to try a new state or action. This process improves the learning effects. The disadvantage is that the convergence time increases exponentially with the duration of exploration.

\subsection{Proximal Policy Optimization (PPO)}
Proximal policy optimization (PPO) is a modified version of the policy gradient method. It used to address the refresh rate problem of policy functions. The algorithms used in the reinforcement learning method can be categorized into dynamic programming (DP), the Monte Carlo method and temporal-difference (TD)~\cite{sutton2018reinforcement}. The policy gradient method uses the Monte Carlo method, wherein samples collected for learning. The advantage of this method is that it can apply to unknown environments. The policy gradient method abandons value functions. Instead, it directly uses rewards to update the policy function, which outputs the probability of taking specific action in an individual state. Thus, it can effectively facilitate decision making for non-discrete actions~\cite{sutton2000policy}.

\subsection{Gramian Angular Field (GAF)}
Foreign exchange prices can be viewed as a time series. Therefore, it is necessary to consider prices over time rather than adopting a set of opening and closing prices as a single state. The paper uses the Gramian Angular Field method, which is a time series coding method consisting of many Gram matrices. 
\begin{figure}[h]
\centerline{\includegraphics[width=6cm]{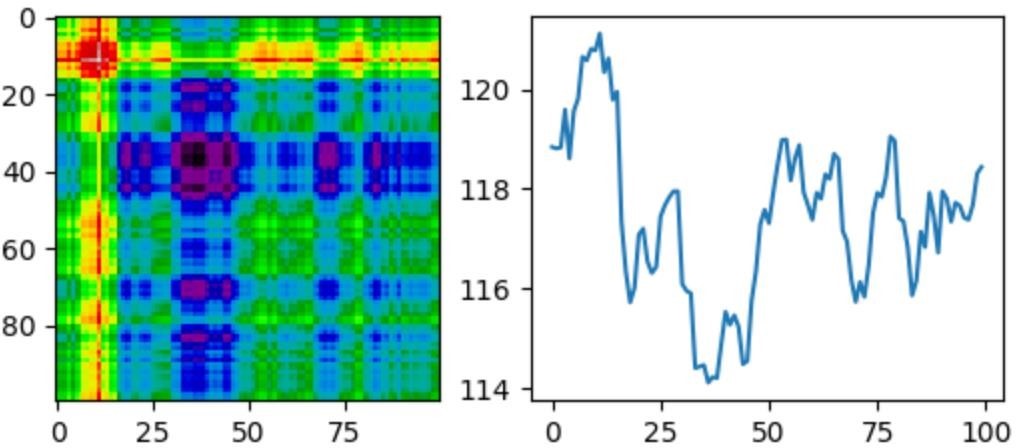}}
\caption{GAP results: Heat-map and numerical line chart.}
\label{fig14}
\end{figure}
A Gram matrix is a useful tool in linear algebra and geometry. It is often used to calculate the linear correlation of a set of vectors~\cite{tsai2019encoding}.
In Fig.~\ref{fig14}, the left image is a heat-map view of the GAF, and the right image is the corresponding numerical line chart. The chart shows high values.

\section{Methodology}
Our reinforcement learning trading system designs as follows:
\begin{enumerate}
\item State Design: States are derived from an agent's observations of the environment. They are used to describe or represent environments. The agents then perform actions corresponding to the perceived state. Therefore, defining the state is key to learning performance.
In the paper, we use the sliding window method on the ``opening price, highest point, lowest point, and closing price'' of the foreign exchange state in units of four hours to obtain sets of 12. After GAF encoding, the data are inputted as a state with dimension $12\times 12\times 4$.
\item Action Design: The purpose of the paper is to optimize the Sure-Fire policy. There are two optimization objectives. The first is to reduce overall buy-ins and the second is to set favorable stop-loss and stop-gain points.
The former is to avoid the lack of funding while the latter is to prevent the fluctuation of prices and unnecessary placement of investments. After pragmatic consideration, three discrete actions are developed:
\begin{itemize}
	\item Upper limit to additional buy-ins after entering the market: $\{1, 2, 3\}$.
    \item First buy or sell: $\{BUY, SELL\}$.
    \item Stop-gain: $\{20, 25, 30\}$.
\end{itemize}
\item Reward Design: Reward is defined as follows:
\[
\textup{Reward}=\textup{Profit}\times\textup{Discount},
\]
where profit is the net income of the current transaction and the discount is
\[
1.0 - 0.1\times\textup{(number of additional buy-ins)}.
\]
This calculation is a variant of the system quality number (SQN). SQNs are one of the indicators used to score trading strategies. The formula is:
\[(\textup{expected profit} / \textup{standard deviation}) \times \]
\[(\textup{square root of the number of transactions}).\]
The discount factor is a variable that decreases concurrently with the increase in transactions. The reason for adding a discount factor to profit is to inform the agent that the reward decreased concurrently with the increase in the number of additional buy-ins. That is, risk increases concurrently with the number of additional buy-ins. Therefore, excessive buy-ins should be avoided.
\end{enumerate}
Oue experiment environment designs as follows:
\begin{enumerate}
\item Experiment Data: we used the EUR/USD, GBP/USD, and AUD/USD exchange data in units of four hours between 1 August 2018 and 31 December 2018 as a reference for the environment design and performance calculation of the system. The period between 1 August 2018 and 30 November 2018 is the training period for the agent. In this period, the agent repeatedly applied the data to learn, eventually obtaining an optimal investment strategy. The period between 1 December 2018 and 31 December 2018 is the agent's performance evaluation period. The agent uses a trading strategy to form decisions. The system accumulates the rewards obtained during the evaluation period, which served as a reference for performance.
\item Trade Environment Settings: EUR/USD, GBP/USD, and AUD/USD are adopted as investment targets. The smallest unit is 1 pip (0.00001 of the price). The default earnings for an action is set to at least 20 pips, which would be greater than the slip price and transaction fee. Therefore, the slip price and transaction fees are not taken into account. Whenever a transaction is tested, the closing price is used as the data point rather than considering whether the transactions are triggered at a high point or low point in the data (that is, of the four values for each day-opening prices, high point, low point, and closing price-the final one is chosen). Time scales smaller than one day rarely contained stop-loss and stop-gain points.
\item Experiment Model Design: a constant agent is added to the experiment model, serving as a reference value for performance. The input of the three different algorithms is a GAF-encoded time series of dimension $12\times 12\times 4$. A CNN is used as the policy network or Q-network.
\end{enumerate}

\section{Experiment Results and Discussions}
\begin{table}[h]
\small
\center
\begin{tabular}{|c|c|c|}
\hline
Model & Algorithm & Currencies\\ \hline
CEU        & Constant  & EUR/USD\\ \hline
DEU        & DQN       & EUR/USD\\ \hline
PEU        & PPO       & EUR/USD   \\ \hline
CGU        & Constant  & GBP/USD\\ \hline
DGU        & DQN       & GBP/USD\\ \hline
PGU        & PPO       & GBP/USD  \\ \hline
CAU        & Constant  & AUD/USD\\ \hline
DAU        & DQN       & EUR/USD\\ \hline
PAU        & PPO       & EUR/USD    \\ \hline
\end{tabular}
\caption{\label{tab1} Model Code Information}
\end{table}
In Table~\ref{DQNPPOResults}, PEU and DEU profits are lower than those of CEU. The accumulated return trajectories for PEU and DEU are almost identical, with the sole differences being the slope. The lines for accumulated returns of PEU and DEU are smoother than CEU. The max drawdown is also much lower, which shows that net profits for PEU and DEU are lower only because no wrong decisions made in the more conservative markup models. The max drawdown of PEU is almost half that of CEU. A more economical risk-orientated strategy for stable returns very possibly discovered for PEU and DEU. The returns for the three are about the same, but the PGU profit factor is much higher than that of the other two, and the PGU max drawdown is the lowest of the three. It can see that PGU using the PPO trade algorithm is the most consistently profitable.
Interestingly enough, the trading decisions made in DGU are almost all reached later than those in PGU. PGU more clearly captured profit maximization properties than DGU during the training period, resulting in a situation in which PGU did not experience the same delays as DGU. PAU made trading decisions that maximized profitability, with every evaluation being far better than that of CAU. Though the max drawdown of PAU is approximately 1.25 times that of DAU, returns are almost 1.5 times those of DAU, while the profit factor remained the highest of the three. Besides the unexpectedly high PGU earnings, there is not much out of the ordinary in the accumulated returns line graphs for DAU and CAU. DAU continually made sound trading decisions, and its max drawdown is the lowest of the three.
\begin{table}[h]
\small
\center
\begin{tabular}{|c|c|c|c|}
\hline
Model Code & Net Profit & Profit Factor & Max Draw-down \\ \hline
CEU        & 753  & 3.67 & -5.18\%  \\ \hline
DEU        & 486  & 3.09 & -3.07\% \\ \hline
PEU        & 615  & 3.82 & -2.92\% \\ \hline
CGU        & 753  & 3.24 & -5.58\%  \\ \hline
DGU        & 717  & 3.10 & -4.63\% \\ \hline
PGU        & 763  & 4.47 & -4.33\% \\ \hline
CAU        & 351  & 1.62 & -14.53\%  \\ \hline
DAU        & 402  & 2.32 & -5.97\% \\ \hline
PAU        & 597  & 2.85 & -7.62\%\\ \hline
\end{tabular}
\caption{\label{DQNPPOResults} Comparing DQN model with PPO model trading performance. Using EUR/USD 4 hours data from 2018/12/01 to 2018/12/31 at 1300 episode.}
\end{table}
\section{Conclusion}
We used a GAF-encoded time series as the state, two different algorithms (DQN and PPO), and three different currency pairs (EUR/USD, GBP/USD, AUD/USD) to create six different trading models. 
From the above comparisons that using PPO’s reinforcement learning algorithm to optimize forex trading strategies is quite feasible, with PPO performance being better than DQN. The results of the various models indicated that favorable investment performance achieved as long as the model can handle complex and random processes, and the state can describe the environment, validating the feasibility of reinforcement learning in the development of trading strategies. We found that the most challenging aspect is the definition of reward.  
\bibliography{article}{}
\bibliographystyle{plain}

\end{document}